\documentclass[letterpaper, 10 pt, conference]{IEEEtran}  % Comment this line out if you need a4paper

\IEEEoverridecommandlockouts                              % This command is only needed if 
\usepackage{graphics} % for pdf, bitmapped graphics files
\usepackage{epsfig} % for postscript graphics files
\usepackage{mathptmx} % assumes new font selection scheme installed
\usepackage{times} % assumes new font selection scheme installed
\usepackage{amsmath} % assumes amsmath package installed
\usepackage{amssymb}  % assumes amsmath package installed

\usepackage[T1]{fontenc}
\usepackage{amsfonts}
\usepackage{booktabs}
\usepackage{siunitx}
\usepackage{caption} 
\usepackage{booktabs}
\usepackage{graphicx}
\usepackage{algorithm, algpseudocode}
\usepackage{float}
\usepackage[percent]{overpic}
\usepackage{caption}
\DeclareGraphicsExtensions{.pdf,.png}

\usepackage[font=small,skip=0pt]{subcaption}

\usepackage{array, tabularx, makecell, booktabs}%
\usepackage{tikz}

\usepackage{geometry}
\usepackage{comment}

\title{\LARGE \bf
DEMO: Arena-Web - A Web-based Development and Benchmarking Platform for Autonomous Navigation Approaches
}

\author{Linh K{\"a}stner$^{1,*}$\thanks{$^{1}$Linh K{\"a}stner, Reyk Carstens, Christopher Liebig, Volodymyr Shcherbyna, Lena Nahrworld, Subhin Lee, and Jens Lambrecht are with the Chair Industry Grade Networks and Clouds, Faculty of Electrical Engineering, and Computer Science,				
		Berlin Institute of Technology, Berlin, Germany
		{\tt\small linhdoan@tu-berlin.de}}\thanks{$^{*}$Contributed equally}, Reyk Carstens$^{1,*}$, Lena Nahrwold$^{1}$, Christopher Liebig$^{1}$,\\ Volodymyr Shcherbyna$^{1}$, Subhin Lee$^{1}$, and Jens Lambrecht$^{1}$
}

\begin{document}

\maketitle
\thispagestyle{empty}
\pagestyle{empty}

% !TeX encoding = utf-8
% !TeX language = en_GB
% !TeX spellcheck = en_GB
% !TeX root = paper.tex

\begin{abstract}

In recent years, mobile robot navigation approaches have become increasingly important due to various application areas ranging from healthcare to warehouse logistics. In particular, Deep Reinforcement Learning approaches have gained popularity for robot navigation but are not easily accessible to non-experts and complex to develop. In recent years, efforts have been made to make these sophisticated approaches accessible to a wider audience. In this paper, we present Arena-Web, a web-based development and evaluation suite for developing, training, and testing DRL-based navigation planners for various robotic platforms and scenarios. The interface is designed to be intuitive and engaging to appeal to non-experts and make the technology accessible to a wider audience. With Arena-Web and its interface, training and developing Deep Reinforcement Learning agents is simplified and made easy without a single line of code. The web-app is free to use and openly available under the link stated in the supplementary materials. 

\end{abstract}
\section{Introduction}
\noindent With recent advances in Deep Reinforcement Learning (DRL) for navigation and motion planning, several research works utilized DRL inside their approach \cite{dugas2020navrep, everett2018motion}. However, developing DRL agents comes along with a number of barriers and difficulties such as difficult training, complex setup, or expensive hardware thus, limited to experts and practitioners in the field. This, makes comparability difficult \cite{xiao2022motion}. However, benchmarking and testing DRL approaches is crucial to compare them against classic approaches and assess their performance and feasibility. Aspirations and efforts to standardize the process of benchmarking were also made in recent years \cite{heiden2021bench, kastner2022arena-bench, tsoi2022sean}. However, the application consists of multiple packages and repositories, which may not be easily accessible and understandable for laypersons.  In addition, they also require an installation and setup process or specified hardware, which increases the barriers for laypeople or new practitioners.
On the other hand, web-based applications have proved to facilitate a more intuitive experience and are more openly available. There have been several efforts to bring robotic applications and assist the user via web apps \cite{marin2002very, siegwart1999interacting, thrun1999minerva}. 

\begin{figure}[!h]
     \centering
     \includegraphics[width=0.95\linewidth]{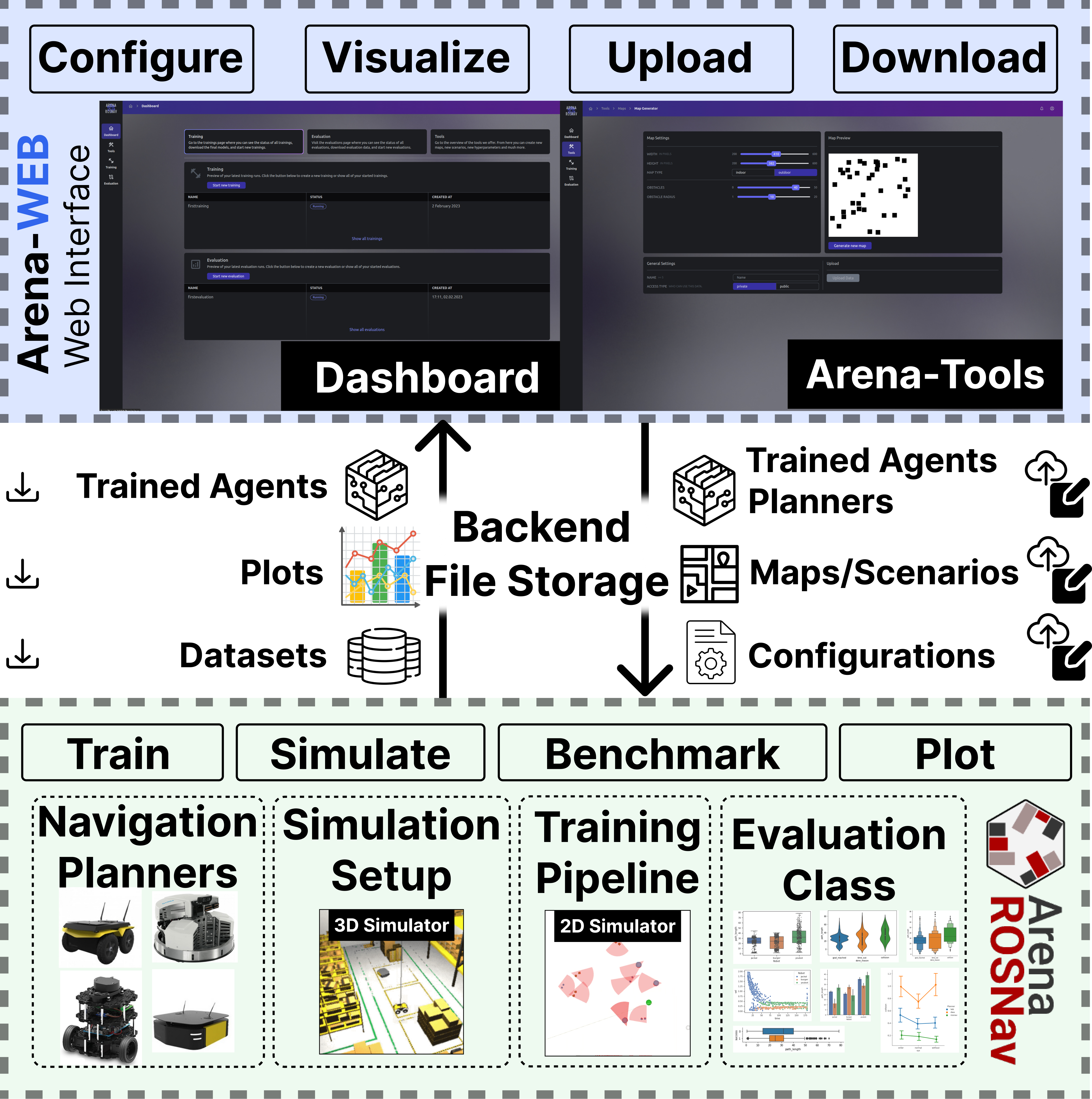}
    \caption{Arena-Web provides a web-based interface to develop, train, and test navigation approaches conveniently on any computer. It provides interfaces to generate a number of utilities such as maps, scnearios, training parameters, and tasks in order to define training and testing runs. The UI is designed in an intuitive and appealing way and the user is guided through a number of predefined steps without sacrificing the high degree of custumization possibilities. The web-app is build on top of our previous work Arena-Rosnav which is run in the backend and contains all the functionalities to run the simulation and generate the data.}
    \label{intro}
\end{figure}

\noindent 
A web-app is easily accessible from all operating systems and does not require specific hardware in order to initiate complex tasks. Furthermore, there is no need to go through complicated installation procedures and face potential errors and issues, which may hamper the motivation to start in this field. Hence, an application that fosters easily accessible development would not only benefit laypeople aspiring to to work with DRL and motion planning approaches but also for expert scientists who are looking for a unified, un-complicated platform to benchmark their approaches and safe time and costs. On that account, we propose Arena-Web, a web-based application, which provides the possibility to develop DRL agents and benchmark these navigation approaches against classic ones using a variety of tools. The user has a high number of possibilities to set and generate different maps, worlds, and scenarios or task modes. All aspects along the pipeline are individually customizable, editable and can be specified by the user according to his or her needs. The user can customize task modes that the agent should be tested on, new worlds and scenarios, neural network architectures, etc. 
Arena-Web is the web interface which combines the functionalities of our previous works Arena-Rosnav \cite{kastner2021towards} and Arena-Bench \cite{kastner2022arena-bench} and adds several more functionalities to them. In particular the web-app is able to do the following: 
train DRL agents on custom maps, with custom reward values, and custom neural network architectures. Evaluation and benchmarking of the approaches in comparison to other learning-based and classic navigation planners. Download of the created models and evaluations files for further use, like plotting of the results or running the models on real robots. Furthermore, there are several editors to create scenarios, neural network architectures, set of hyperparameters, maps, and custom rewards. All functions are embedded into an intuitive and appealing web application for improved user experience, understanding, and accessibility. The main contributions of this work are the following:
\begin{itemize}
    \item To the best of our knowledge, the first web-based DRL development and testing platform that lets the user create, train, and benchmark DRL agents with a variety of classic and learning-based navigation approaches. the web app is built on top of our previous works arena-bench, which is modular and can be extended with other planners and approaches.
    \item Possibility to customize the whole training pipeline from customized network architectures to hyperparameters, reward functions, and scenarios. 
    \item Possibility to customize the whole evaluation pipeline from customized scenarios, task modes, and planners. A large number of qualitative and quantitative evaluation metrics and plotting functionalities are available.

\end{itemize}

\noindent The paper is structured as follows. Sec. II begins with related works. Subsequently, the methodology is presented in Sec. III. Sec. IV presents exemplary use cases that can be executed with the app. Finally, Sec. V will provides a conclusion and future works.

% \newgeometry{top=0.75in,bottom=0.75in,right=0.75in,left=0.75in}
\section{Related Works}
\noindent DRL-based navigation approaches proved to be a promising alternative that has been successfully applied in various robotic applications with remarkable results. Thus, various research works incorporated DRL into their navigation systems \cite{dugas2020navrep, everett2018motion, chen2019crowd, faust2018prm, chen2017socially, chiang2019learning, guldenring2020learning, kastner2021towards, kastner2021connecting}.
\\\noindent
However, most of these research works require at times tedious installation and setups in order to test and validate the approaches. Oftentimes, depreciated versions require manual adjustments and bug fixing, which consumes cost and time. Furthermore, standardized platforms to benchmark DRL-based approaches with classic ones side-by-side are rare and limited \cite{xiao2022motion}, \cite{kastner2022arena-bench}. There exist a number of platforms that aspire to provide a platform for benchmarking and developing these approaches \cite{heiden2021bench, kastner2022arena-bench, tsoi2020sean, tsoi2022sean}. However, they also require an installation and setup process or specified hardware, which is not intuitive or possible for laypersons or new practitioners. 
On the other hand, web-based applications have proved to facilitate a more intuitive experience and are more openly available to a larger audience.
\\\noindent
There has been a number of efforts to provide services from the robot operating system as web applications. For the past two decades already efforts has been made to combine robotics with web based applications. In 2000, Schulz et al. \cite{schulz2000web} proposed web based interfaces to remotely control robots via the internet. The researchers provided a web-based user interface in which the user could select a map and navigate a robot to a desired point within the map. Similarly, Simmons et al. proposed an extended and improved version with more functionalities and visualizations. Other web-based platforms for robot control and interaction include \cite{marin2002very, siegwart1999interacting, koch2008universal, thrun1999minerva}
\\\noindent
Using web interfaces also prove practical for an improved human-robot-collaboration and -interaction since web services are easier to access and more openly available than command line tools or development setups. Thus, more recent robotic web-based applications range from robot assistance in healthcare \cite{bhattacharjee2020more, di2014web, cocsar2020enrichme}, manufacturing \cite{perzylo2019smerobotics},  guiding robots \cite{del2019lindsey, bavcik2020phollower} or educational purposes \cite{canas2020ros,hormaza2019line, faina2020evobot, bejarano2019implementing}
\\\noindent
Furthermore, with the advent of the robot operating system (ROS) \cite{quigley2009ros}, tools have been proposed to combine ROS with web. Open-source libraries such as ros.js proposed by Osentoski et al. \cite{osentoski2011robots} or the Robot Web Tools proposed by Toris et al. \cite{toris2015robot}, which provides major ROS functionalities via rosbridge as a web app, contribute to foster web-based robotic development and usage to broaden the development audience. More recently, tools such as Webviz \cite{webviz} or Foxglove \cite{foxglove} aspire to shift ROS based visualization and monitoring entirely to the web.
\\\noindent
On that account, and given the limitations of a unified evaluation and benchmarking platform, this paper proposes Arena-Web to facilitate the development of DRL-based approaches and evaluation against other learning-based and classic navigation planners. The platform is made available entirely via the web application to facilitate swift development and intuitive usage thus, accelerating the development cycle. The tool aspires to open DRL development and benchmarking of robot navigation approaches to a wider audience.

\begin{figure}[!h]
    \centering
    \includegraphics[width=0.99\linewidth]{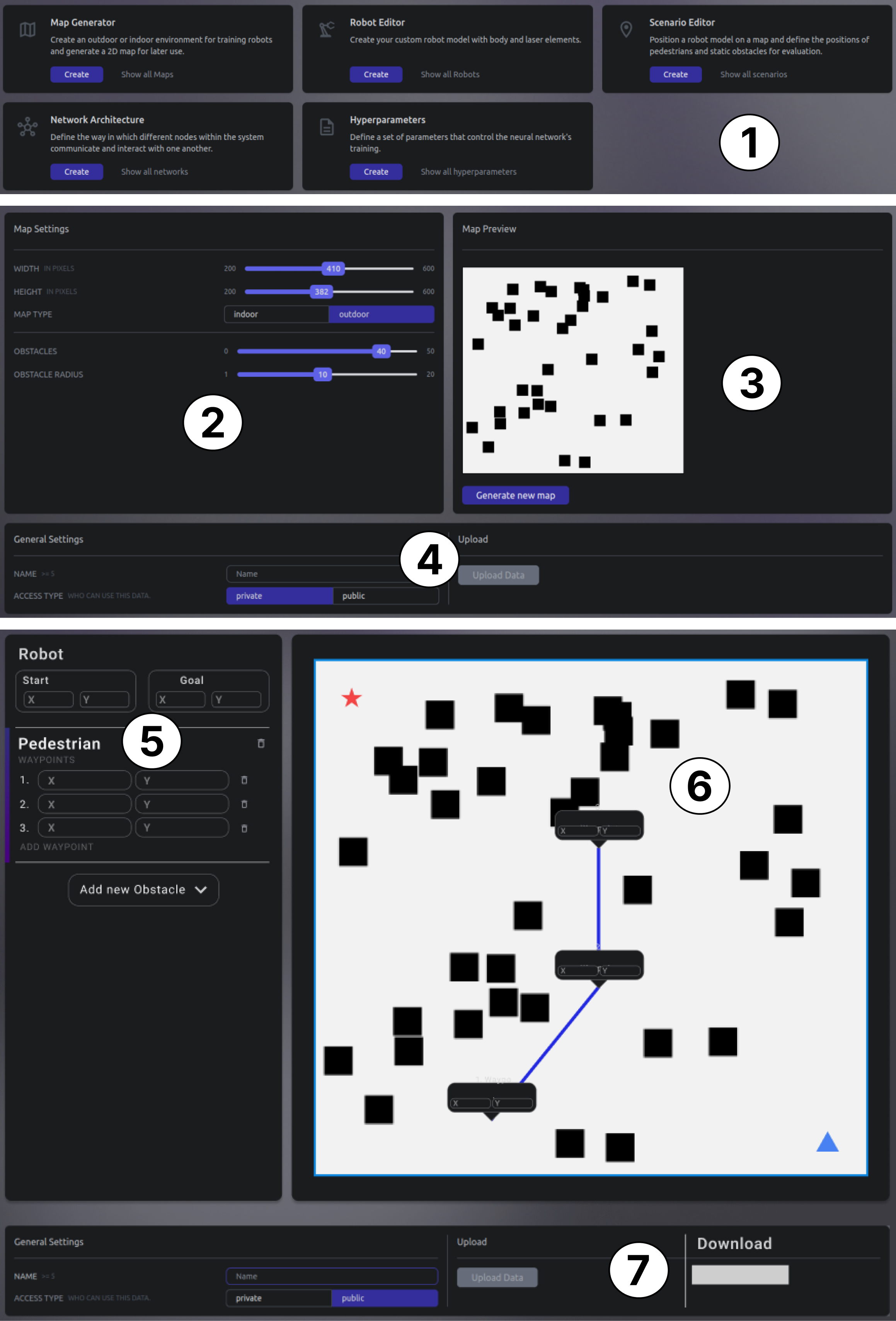}
     \caption{We provide a number of editors/generators to generate utilities such as an occupancy map, scenarios with predefined obstacle numbers and behavior, training pipeline editors to generate new neural network architectures, select training hyperparameters, reward systems, etc. (1) The dashboard to select the tool, (2) Parameters regarding the map can be adjusted using sliders, (3) the map is previewed live so that the user can directly observe changes, (4) it is also possible to upload existing maps, (5) in the scenario editor the user can input positions of the robot and obstacles by drag-and-drop, the values will be displayed in the sidebar, (6) the scenario is previewed to the user, (7) the user can upload existing scenarios. }
     \label{fig:tools-screens}
\end{figure}

\section{Functionalities}
\label{sec:functionalities}
\noindent Arena-Web is the web interface that combines the functionalities of our previous works Arena-Rosnav \cite{kastner2021towards} and Arena-Bench \cite{kastner2022arena-bench}, adds several more functionalities to them, and serves as an openly accessible user interface. The main tasks of the application are divided into three categories: Tools, Training, and Evaluation. This section should provide an overview of the current functionalities of Arena-Web. It is to be noted, that more functionalities are constantly being added to the application.
\subsection{Tools}
\noindent
We provide tools to generate several important utilities necessary to conduct training or evaluation tasks. An overview of all currently available tools can be seen in Fig. \ref{fig:tools-screens}.
\\\\\noindent
\textbf{Map Generator:} 
    \noindent The map generator is able to generate 2D worlds for indoor and outdoor environments. The user can specify various parameters such as the height and width of the map, the amount and size of static obstacles, the width of passages between room and the size and amount of rooms in indoor maps. The map generator will generate an occupancy grid map.
    % For indoor maps this means more narrow passages while for outdoor maps more static obstacles are spawned. 
    % In case of a 2D map, a ros occupancy grid map will be saved and for 3D a gazebo world file is saved to the filesystem for further usage.
%\\\noindent
\\\\\noindent
\textbf{Scenario Generator:}
    \noindent The scenario editor allows the user to select a map and place dynamic obstacles inside, define their velocities, and define their behavior by setting waypoints. The robot's start and endpoint is also to be set. The user can edit all positions by drag-and-drop, which simplifies and improves the user experience. The scenario can later be used in the evaluation phase to offer the same environments for different planners and allow a consistent way of comparison.
    % TODO Reset count timeout time
    
    % the user can then specify the behavior of those dynamic obstacles by dragging and dropping. The user can specify the obstacle velocity, their waypoints, their movement patterns and behavior. Furthermore, social states of the obstacles can also be set.  A scenario.yaml file will be created which can be used for the task generator which can read this yaml file and create the respective scenario for either training or evaluation purposes.

% \textbf{The robot editor}.
%     Using the robot editor, the user can create a new robot model for training. this is especially helpful for the 2D environment since there exist few 2D models and the user can specify individual shapes. currently only 2D model creation for training in flatland 2d simulator is possible. the user does this by dragging and dropping and adding functions to the editor which generates a config yaml file. The pluggins such as laser scan or diff drive pluggin can be added. the yaml file required for flatland to recognize the object is then produced and saved to the filesystem. 
% \\\noindent
\noindent
\textbf{Network Generator:}
    Using the network generator, the user can create an individual neural network architecture for the training of DRL agents. The user can specify the in- and output-size of the different layers and is able to add and remove these layers. Currently, fully connected, convolutional, and Relu layers are possible, but our modular configuration makes it easy to add more layer types in near future. 
    % the network is saved as a py file into the file system for futther usage with the training pipeline.
\\\\\noindent
\textbf{Reward Generator:}
    Since DRL is used to train the agent, rewards have to be defined for specific actions. Adapting the rewards can highly influence the outcome of the training. Therefore, with the reward generator, we offer a way for the user to adapt all rewards to their specific needs.
\\\\\noindent
\textbf{Hyperparamters Generator:}
    Similar to the rewards, the hyperparameters are an essential and important component of the training process. Since setting up hyperparameters is oftentimes a trial-and-error process, with the hyperparameters generator we give the user the possibility to create their own set and try different values.
\begin{figure*}[!h]
    \centering
    \includegraphics[width=0.99\linewidth]{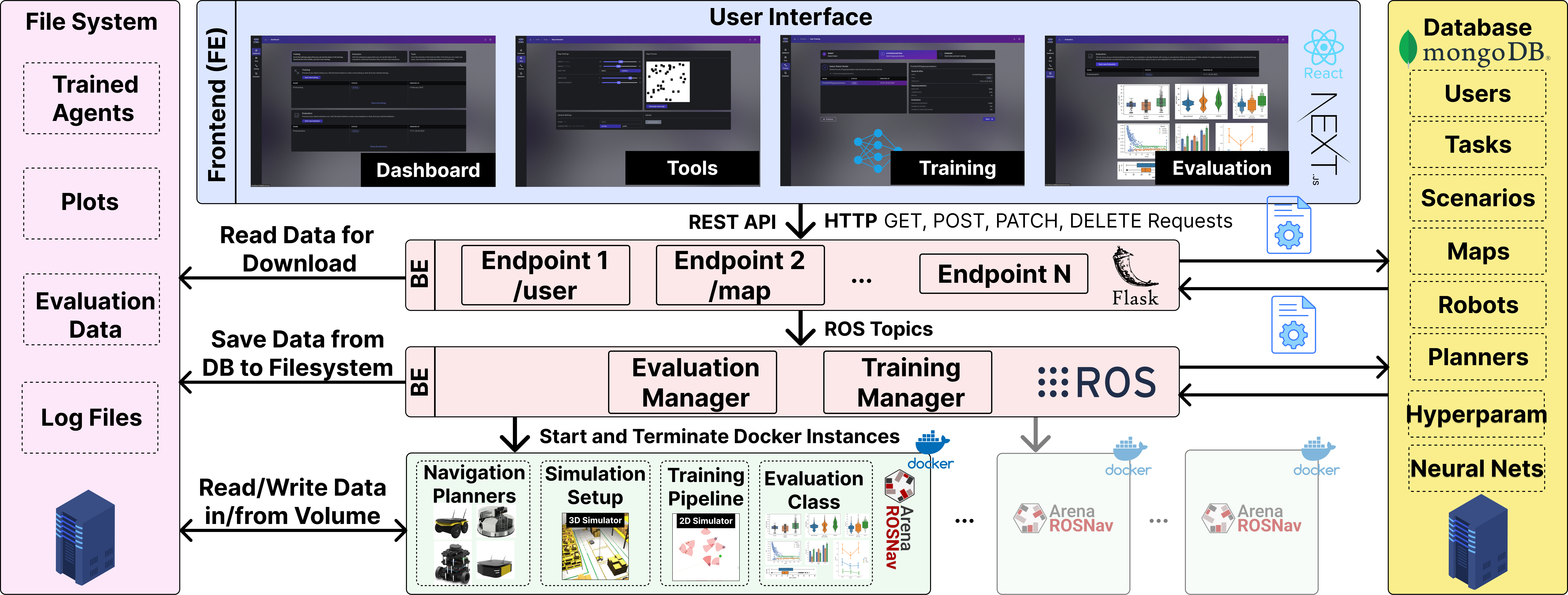}
    \caption{System Design of Arena-Web. The frontend (FE) provides user interfaces to customize and setup training and evaluation runs and construct necessary utilities such as maps and scenarios. The web-app is build on top of Arena-Rosnav (see Fig. \ref{fig:arena-rosnav} for the system design of Arena-Rosnav), which provides tools to train and benchmark DRL planners against other learning-based or classic approaches and plot the results on different metrics. Therefore, the FE is communicating with the backend (BE) via a REST API and endpoints and an additional ROS layer consisting of two ROS nodes to create and manage the Arena-Rosnav dockers in the BE. The generated files are stored within a MongoDB database as well as a file system for user generated data. The backend and databases are run on our GPU servers.}
    \label{fig:system}
\end{figure*}

\subsection{Training}
% \textbf{Training} 
\noindent One major task of our web-app is the ability to train a new DRL model for a robot. Therefore, we offer a preset of 10 robots to choose from. The training can be done on predefined or custom-created network architectures and maps. Furthermore, the user is able to specify a variety of hyperparameters such as the learning rate, the batch size, the evaluation frequency. While the training is running on the servers, the user is always able to directly download the current best model and see the log output to monitor the process. When the training has finished, the user will be notified. Since the training is always dependent on a robot, we offer a large variety of pre-available robots, covering all major robot kinematics. This includes the Turtlebot3, Robotino Festo, Carobot4, AGV Ota, Ridgeback, Kuka Youbot, Clearapath Jackal, Clearapath Husky, Dingo, TiaGo.
    
    % and the training setup which was created using the training editors or also uploaded. the currently available robots are listed in table x. The user can set several important hyperparameters such as the learning rate, etc. and the neural network the algorithm, the evaluation counters, step size etc. 

\subsection{Evaluation}
\noindent The evaluation task can be utilized to evaluate the trained agent or other existing planners within the Arena-Rosnav suite. All available planners are listed in Table \ref{tab:planners}. We provide a large number of state-of-the-art learning-based as well as classic planners. By selecting predefined scenarios for the evaluation, the user can make sure to offer the same environment for all of the evaluations and have a consistent comparison. At all times, the user is allowed to download the current evaluation data while the status of the evaluation run can be observed using the log files. Once the evaluation run is finished, the user can download the resulting .csv files to plot the results locally with our evaluation package. Users have the possibility to create their own plot declaration files, which defines what plots have to be created and what data to use for them. This allows even inexperienced users to quickly plot their results. Currently, we are working on integrating the plotting directly into the web app for an improved user experience.

    % the user can set different evalution tasks such as evaluting on a specific scenario which was created or uploaded or on randomly generated maps, 
    % the number of episode each scenario should be run on, the timeout time, etc. once all parameters are set, the evaluation will trigger the simulation run either on a 2d simulator or on the 3D gazebo. the user will be shown a progress bar indicating the status of the evaluation run. Once the evaluation tasks are done, the respective csv files with all recorded metrics are saved into the file system from which the user can donwload the data. Subsequently, the data can be used to plot individually or our plotting class can handle the plotting and display the results directly within the webapplication.

\subsection{Available Robots and Planners}
\noindent There are a variety of already pre-available robots covering all major robot kinematics. These include the Turtlebot3, Robotino Festo, Carobot4, AGV Ota, Ridgeback, Kuka Youbot, Clearapath jackal, Clearapath Husky, Dingo, TiaGo.
Furthermore, Table \ref{tab:planners} lists all available planners.
\begin{table}[htbp]

\centering
\setlength{\tabcolsep}{12pt}
\renewcommand{\arraystretch}{1}
\setlength{\tabcolsep}{.5\tabcolsep}
\footnotesize
\caption{Available Navigation Planners}
\begin{tabular}{@{}lccccc@{}}\toprule
Classic & Hybrid & Learning-based   \\ 
\midrule
TEB \cite{rosmann2015timed} &  Applr \cite{xiao2020appld} & ROSNavRL \cite{kastner2021towards}  \\
DWA \cite{khatib1986real} &  LfLH \cite{xiao2022motion} & RLCA \cite{long2018towards}  \\
MPC \cite{rosmann2019time} &  Dragon & Crowdnav \cite{chen2019crowd}  \\
Cohan \cite{singamaneni2021human} &  TRAIL & SARL \cite{li2019sarl}  \\
 &   & Arena \cite{kastner2020deep}  \\
  &   & CADRL \cite{everett2018motion}  \\
\bottomrule
\end{tabular}
\label{tab:planners}
\end{table}

\begin{figure*}[!h]
     \centering
     \includegraphics[width=0.99\linewidth]{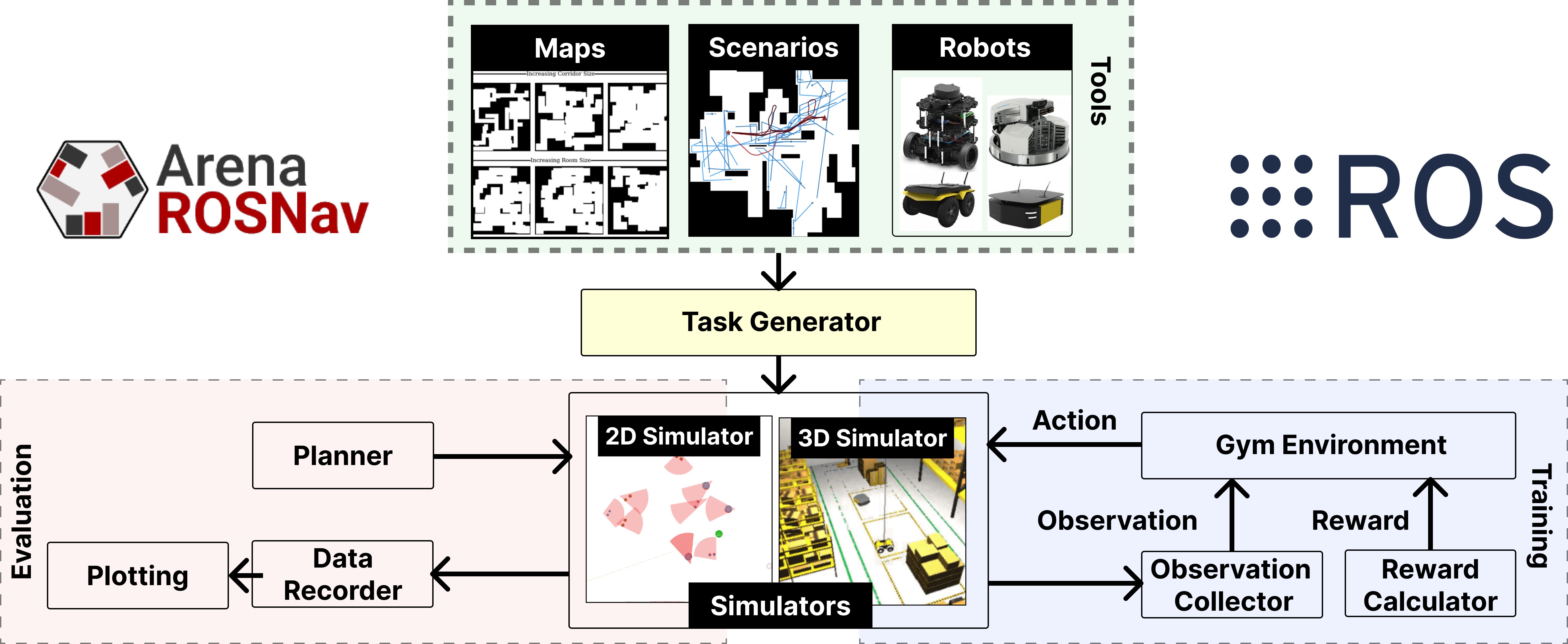}
     \caption{System design of Arena-Rosnav. It provides tools to develop, train, and evaluate DRL planners on different environments and scenarios, especially in highly dynamic environments. Therefore, it includes multiple simulation environments such as Flatland (2D), Gazebo (3D), or Unity (3D). Note that currently the user can only choose simulation within the 2D environment. We are currently working on enabling 3D environments to be choosen. The training pipeline conducts a DRL training, which can be customized by the user. Subsequently, the trained agent can be benchmarked against other navigation approaches. Arena-Rosnav provides a suite of different state-of-the-art planners listed in Table \ref{tab:planners} and an evaluation pipeline including a data recorder and a plotting class. The data recorder records all necessary data during the evaluation run and the plotting class can utilize these files to plot the results on a number of navigational metrics such as navigational efficiency, saftey, and smoothness.}
     \label{fig:arena-rosnav}
\end{figure*}

\section{Methodology}
\subsection{System Design Overview}
\noindent
A schematic overview of the whole system is shown in Figure \ref{fig:system}. The user interacts with the user interface in the frontend, which is implemented in React.js. The frontend It can be segregated into three main segments, the tools, the training, and the evaluation. The frontend communicates with the backend over a REST API. The backend is written in flask and offers multiple different endpoints for all functionalities of the web app. For persistent and efficient storage, a MongoDB database is used, which is chosen over a relational database because of its possibility to handle unstructured data. When starting new training and evaluations, Arena-Rosnav is run as a Docker on the server to process the task. The system design of Arena-Rosnav is illustrated in Fig. \ref{fig:arena-rosnav}. Since the process should be customizable by the user's created tools and the output and created files should also be available to the user, parts of the server's file system are virtually loaded as volumes into the docker to save and load files from outside of the docker. For an extra layer of security and easier managing of the different docker containers, a small layer of ROS-Nodes is implemented between the flask backend and dockers.

\subsection{Frontend}
\noindent
The Frontend offers user interfaces for starting new training and evaluation runs as well as creating new utility files for use in either training or evaluation with the tools as described in \ref{sec:functionalities}. For security and management reasons, the whole frontend is only available to already registered users, which are logged in. This allows to always relate user-created data and tasks to the specific user and therefore keep the whole interaction private. We have decided to uncouple the direct connection between utility files and specific tasks so that created files can be used in combination with other utilities and for different and multiple tasks. This allows for modularity and experiment with even less work.
\\\\\noindent
In the frontend, the different tools are used to create the requisites for customized training or evaluations. As such, a map-, scenario-, network architecture-, hyperparameter-, and reward generator were implemented, which have already been presented in section \ref{sec:functionalities}. We have decided to make it possible to create all utility files used in training and evaluation separately from specific tasks in order to reuse them for different and multiple runs instead of having to repeatedly creating them again for different training or evaluation runs.

\subsection{Backend}
\noindent
The backend offers a variety of different REST Endpoints to compute every functionality that is possible in the frontend. The backend uses a MongoDB database for persistent and efficient storage of all data created using the tools. All created utility files, which are created with the tools, as well as training and evaluation infos are stored inside of the database for easier manipulation and consistent access from all parts of the backend. Beside the flask backend the server runs a small topology of ros nodes to manage docker containers, in which the training and evaluation tasks are run. Therefore, when starting new tasks the flask backend tells the ros topology to start the defined task with specified configurations included in the request. The Ros Nodes will then create the necessary utility files and store them in the filesystem of the server. Subsequently, a new docker container for the task is started and the utility files are loaded into the docker using docker volumes. To allow the user to download created data, the output directories of files in the docker are also loaded as volumes to be visible in the server's file system and be usable by the backend.
\\\\\noindent
Using the docker volumes not only allows to use of created data later on but also allows the use of the same docker for different tasks and with different utility files. Using this extra layer of ros nodes has multiple advantages. At first, it offers another layer of security because it is not attached to the backend and therefore, failures in or attacks on the flask backend will not affect the docker containers. Furthermore, it allows to be independent by the flask backend and makes possible further integration into other applications way easier. And at last, it leads to a much quicker response to for the user because the endpoint itself only has send a single rosmessage instead storing all files and starting the docker on its own. This way the computation intense can be done asynchronously.
\\\\\noindent
In comparison to the previous training and evaluation capabilities of Arena-Bench, a number improvements have been made. To allow the modularity with which different neural network architectures, rewards, maps, and scenarios can be used, we defined completely new formats to store the data and integrated efficient parsers to read and use the data in \textit{Arena-Bench}. Additionally all of the data is validated in the frontend and in the backend to detect possible errors as soon as possible and reduce errors.

\begin{figure*}[!h]
 	\centering
 	\includegraphics[width=\textwidth]{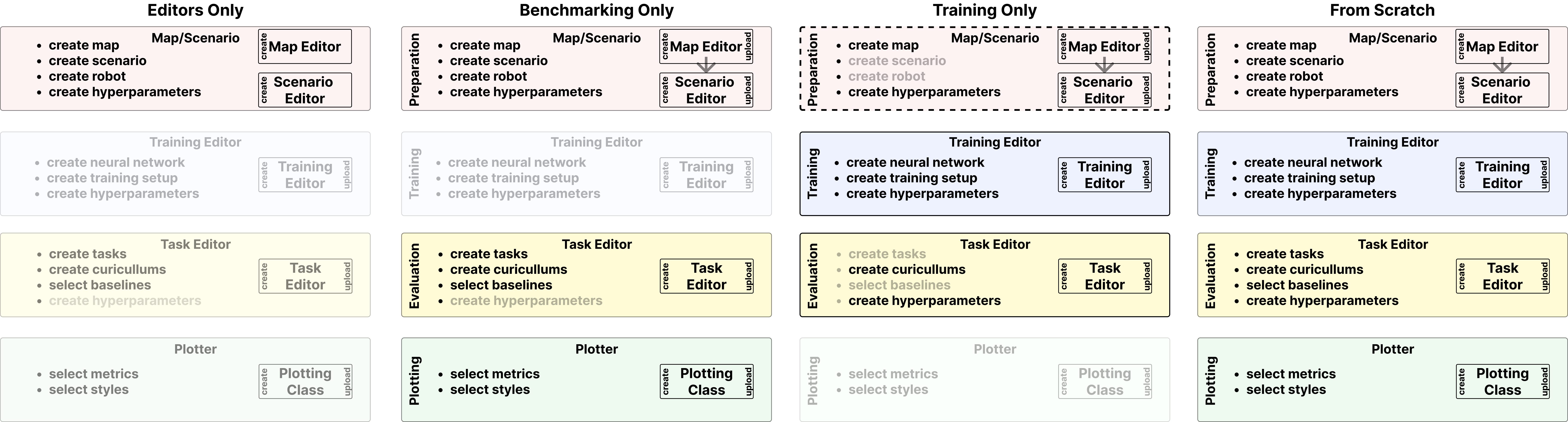}
 	\caption{Exemplary workflows and their necessary modules (blended in for the respective use case). Using Arena-Web can serve different use cases and needs. The user can utilize the variety of editors to create ROS occupancy grid maps or different scenarios using the intuitive interface to simplify the generation process of these files for working with a local instance of Arena-Rosnav. The user can also use the web-app to initiate a DRL training from scratch and benchmark the resulting agent against several planners. Other possibilities include to utilize the app for conducting only a training using existing maps and agents or conducting only evaluation tasks with preexisting agents, planners, and scenarios without accessing the development PC. This could be useful e.g. for teams to conduct multiple evaluation runs without requiring the exact PC setup from anywhere. We are currently working on a mobile version to also facilitate starting or monitoring these tasks from anywhere, which makes the development process as well as collaborations more flexible and efficient.}
 	\label{fig:workflows}
\end{figure*}

\section{Exemplary Use Case Workflows}
\noindent In this chapter, we illustrate how the application can be utilized for different use cases and needs. In particular, there are three main use cases, which are covered using the web-app: creating new utility files like maps, scenarios, rewards, and hyperparamters, training DRL agents, and evaluating existing planners or trained DRL agents, which have been created with the web-app by the user. For each of these use cases, we define an example use case and show the user journey through the web-app and specify the required tools and interactions. The exemplary use cases and respective required tools are illustrated in Fig. \ref{fig:workflows}. The example use cases are as follows:

\begin{itemize}
    \item Creation of a new scenario
    \item Starting a new training and downloading the model.
    \item Starting a new evaluation with a trained DRL agent, downloading the evaluation data and plot the results. 
\end{itemize}
\noindent
A demo video showcasing all of the functionalities is given in the supplementary materials.

% \noindent In this chapter, we illustrate how the application can be utilized for different use cases and needs. In particular, there are four main use cases which are covered using the web application: creating maps or scenarios for ROS for self-use, training DRL agents and evaluating them, or benchmark and evaluate navigation planners.  For each of these use cases, we specify the use case, the required tools, the worflow which the user has to go through, and the end result. The neccessary modules required for each of these tasks are illustrated in Fig. \ref{fig:workflows}.

% \subsection{Creation of ROS Maps and Scenarios}

\subsection{Creation of a New Scenario}
\noindent
The scenario serves as a fixed environment in the evaluation for better comparison of different planner approaches. It specifies the start and goal position of a robot as well as different dynamic obstacles like pedestrians. Furthermore, the exact behavior of the obstacles can be specified by waypoints and the velocity of the obstacles can be set. In order to create a new scenario, the user has to select a map, which can be created by the map generator tool. 
\\\noindent
After logging in, the user is on the Dashboard, which shows a small overview over the latest tasks. From here, the user can directly jump to the tool selection and start creating a new map. After selecting the parameters, the map can be uploaded. Afterwards, the user jumps back to the tool selection and selects to create a new scenario. Here, the user will be asked to select one of the maps the user has access to. Subsequently, the user will be directed to the actual scenario generator page where the map is visualized on the right side. On the left side the user can set the different options in a list and has the possibility to drag and drop the goal of the robot and start position onto the map or directly input X and Y coordinates. Furthermore, the user will find a dropdown menu beneath it from which the type of a new dynamic obstacle, which is then added to the scenario, can be selected. For each obstacle, the velocity can be changed by an input field and new waypoints can be added, which can be changed by dragging them in the map or directly passing in the coordinate values. After finishing the scenario the user is prompted to give the scenario a name and select whether other users are allowed to see and use the created scenario. After hitting upload the process is finished and the scenario can either be downloaded directly or be used for a new evaluation.

% Using the editors, the user can create ros maps and scenarios. Therefore, the just the ros tools need to be employed. Using the user interface provides an intuitive interface to create maps. using these maps a scenario can be generated. it is to be noted that these scenarios are specificially designed to work with the task generator class of arena rosnav. that is, it can read out and interpret the resulting configuration file and create the respective scenario. the user also has the possibility to upload maps and scenarios which have been crerated previously either using the local instance of arena-rosnav or another tool. 

\subsection{Starting a New Training Task}
\label{sec:evaluation:start_training}
\noindent The training can be started with a variety of customization possibilities, since the network architecture, hyperparameters, map, and rewards can be selected and specified. Due to the high number of possibilities and combinations, which could make the user experience overwhelming, we designed an interactive and straightforward pipeline consisting of four steps for selecting the different utilities and starting the training. For all steps, we already offer a variety of openly accessible files to use, and have defined a pre-selected default values. 
\subsubsection{\textbf{Selecting the Map and Robot}}
\noindent After entering the app and being located at the dashboard, the user can directly jump to the page where a new training is created. There, the user is prompted to select the map to train on. Either the user selects a map from high number of pre-existing maps or from self-generated or uploaded ones. After selecting a map, the user will need to select a robot on which the trained model should run. Based on this decision, the neural network architecture step will consider the observation and action space depending on the robot since all robots have different observation spaces.
\subsubsection{\textbf{Creating Sets of Hyperparameters}}
\begin{figure}[!h ]
     \centering
     \includegraphics[width=0.99\linewidth]{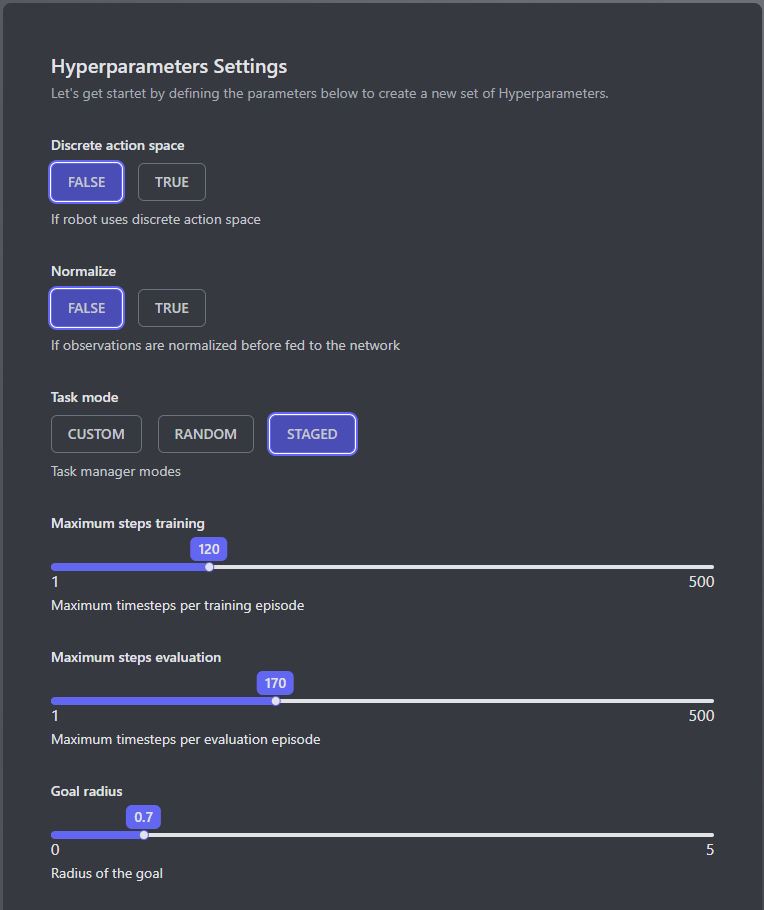}
     \caption{Hyperparameter setting for the training tasks. The user can choose from a number parameters such as the task mode, training step size, evaluation step size, etc.}
     \label{fig:param-settings}
\end{figure}

\noindent
As stated beforehand, sets of hyperparameters necessary for training should be configured in the UI. The UI is depicted in Fig. \ref{fig:param-settings}).

\noindent 
The first step is “Settings”, where the user is able to set each parameter to the desired value (see Fig. \ref{fig:param-settings}). As the hyperparameter generation should be more intuitive than simply putting numbers or strings into an input field, buttons and sliders are displayed instead. A small text below each hyperparameter describes the usage of such. The usage of sliders and buttons also avoids parameters being set to values out of bounds. While parameters with buttons as options can only be set to the displayed values, sliders have a range of what the parameters can be set to. Initially before user configuration, all hyperparameters are set to predefined values as displayed in Fig. \ref{fig:param-settings}.
\\\\\noindent 
After configuring the hyperparameters, choosing their access type is the user's next step in "General Information" either as public, to enable public usage, or private. Based on this decision, other users might see the created hyperparameters. In this step, the user is also asked to name the created hyperparameters. This is important to avoid confusion on the user side. Finally, the user is given an overview of the chosen settings and access type and can submit hyperparameters. 
\subsubsection{\textbf{Creating Neural Network Architectures}}
\noindent
For simplicity, the process of creating custom NN architectures is similar to creating a set of hyperparameters. The NN architecture generator is structured in analogous three steps. The first step "Architecture" guides the user through the creation of the network architecture. This part, however, is more complex than in the hyperparameter generator. 
\begin{figure}[!h ]
    \centering
     \includegraphics[width=0.99\linewidth]{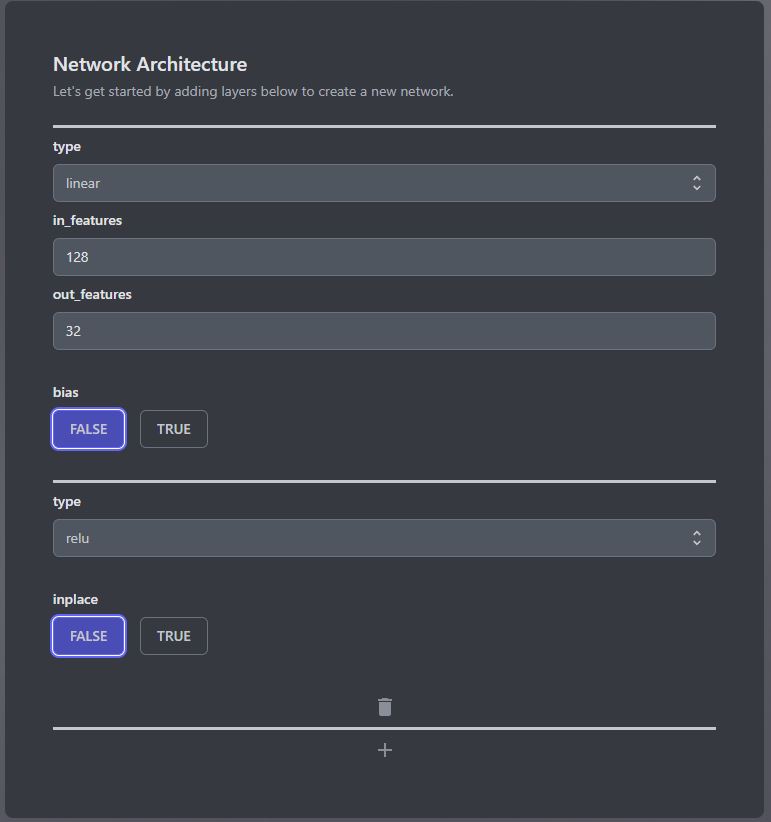}
     \caption{The user interface to define the neural network architecture of the DRL agent. The user can define the type of layers, the input- and output size, or the activation function. The architecture will be visualized for improved understanding. Furthermore, based on the robot, there may be error messages if the in- and output size do not match with the robots specifications.}
     \label{fig:nn-arch}
 \end{figure}
 \noindent
The NN architecture configured at any time is saved in an array. The UI can be seen in Fig. \ref{fig:nn-arch}. By clicking on an add button, the user extends the architecture with a new module and thereby adds another module to the existing array. The last module added can also be deleted using the trash icon. A module consists of a type and the parameters needed for the type of module. In this example, it can be seen that linear layers have the parameters input features, output features and bias. When adding a new module, initially, it is a linear layer but its type can be changed to all available modules. Once finished with the architecture, the user has the possibility to make it public or keep it private and set its name in the “General Information” step. In the “Summary” step the user can see the types of modules combined and the access type once more before submitting it to the backend.
\subsubsection{\textbf{Specifying the Reward System and Starting a Training}}
\noindent
In the final step, the user should specify the reward system by specifying values for a number of preset penalties and rewards using the reward generator. A default reward system is already set so that the user can also skip this step and start the training immediately. To start the training once all necessary input was provided, the user can submit the training task and a JSON object including the name of the training and the ids of the selected robot, network architecture, and hyperparameters will be sent to the backend. The backend will start \textit{Arena-bench} and the training based on the selected configurations. After starting the training the user will be redirected to a page where the log output of the training, and the current best model can be downloaded.

\subsection{Starting a New Evaluation Task}
\begin{figure}[!h ]
     \centering
     \includegraphics[width=0.99\linewidth]{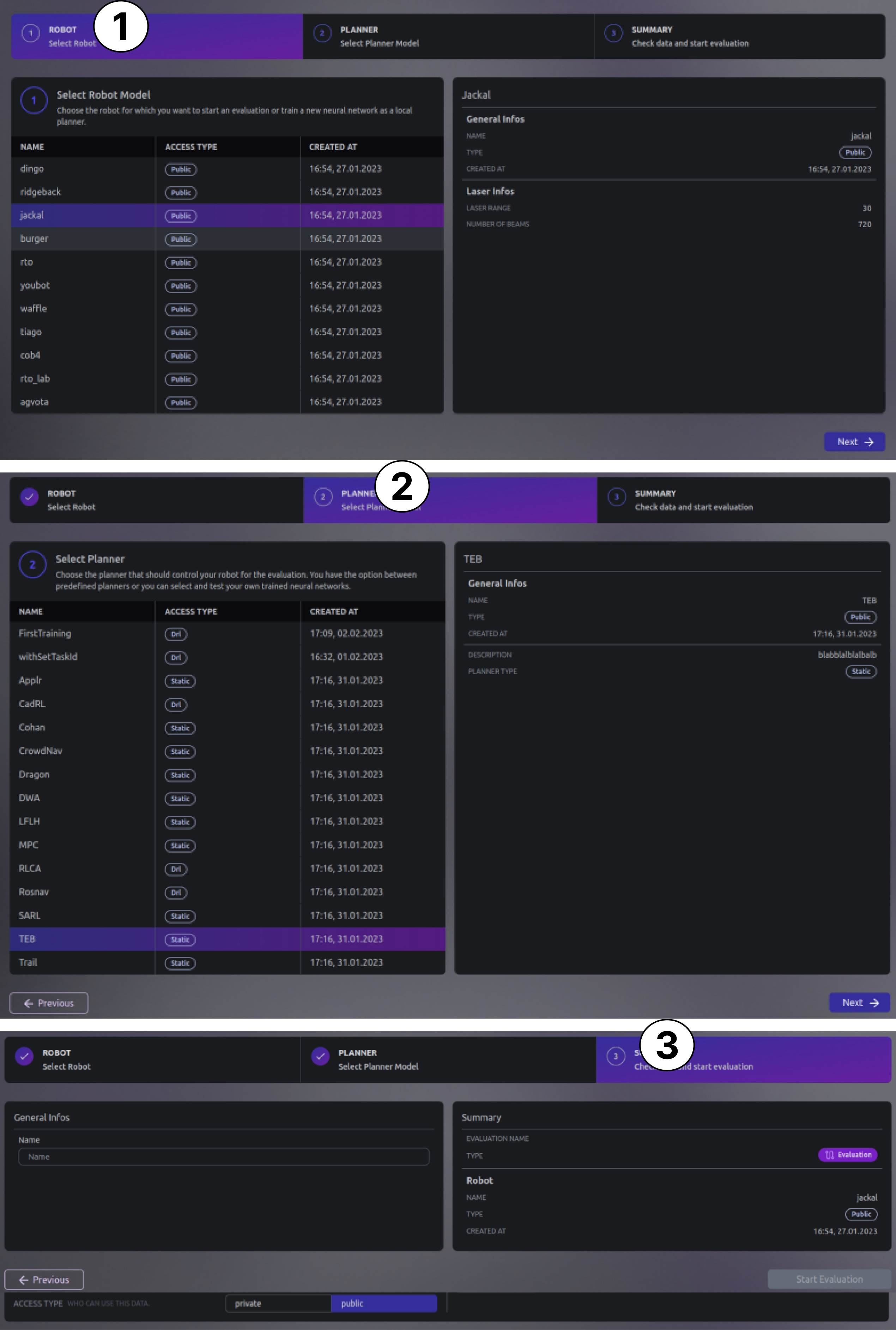}

     \caption{Exemplary pipeline when conducting an evaluation run. It consistist of three steps: (1) choosing the robot model, (2) choosing the planner, and (3) choosing the evaluation tasks.}
     \label{fig:eval-screens}
 \end{figure}

\begin{figure}[!h]
    \centering
    \includegraphics[width=0.99\linewidth]{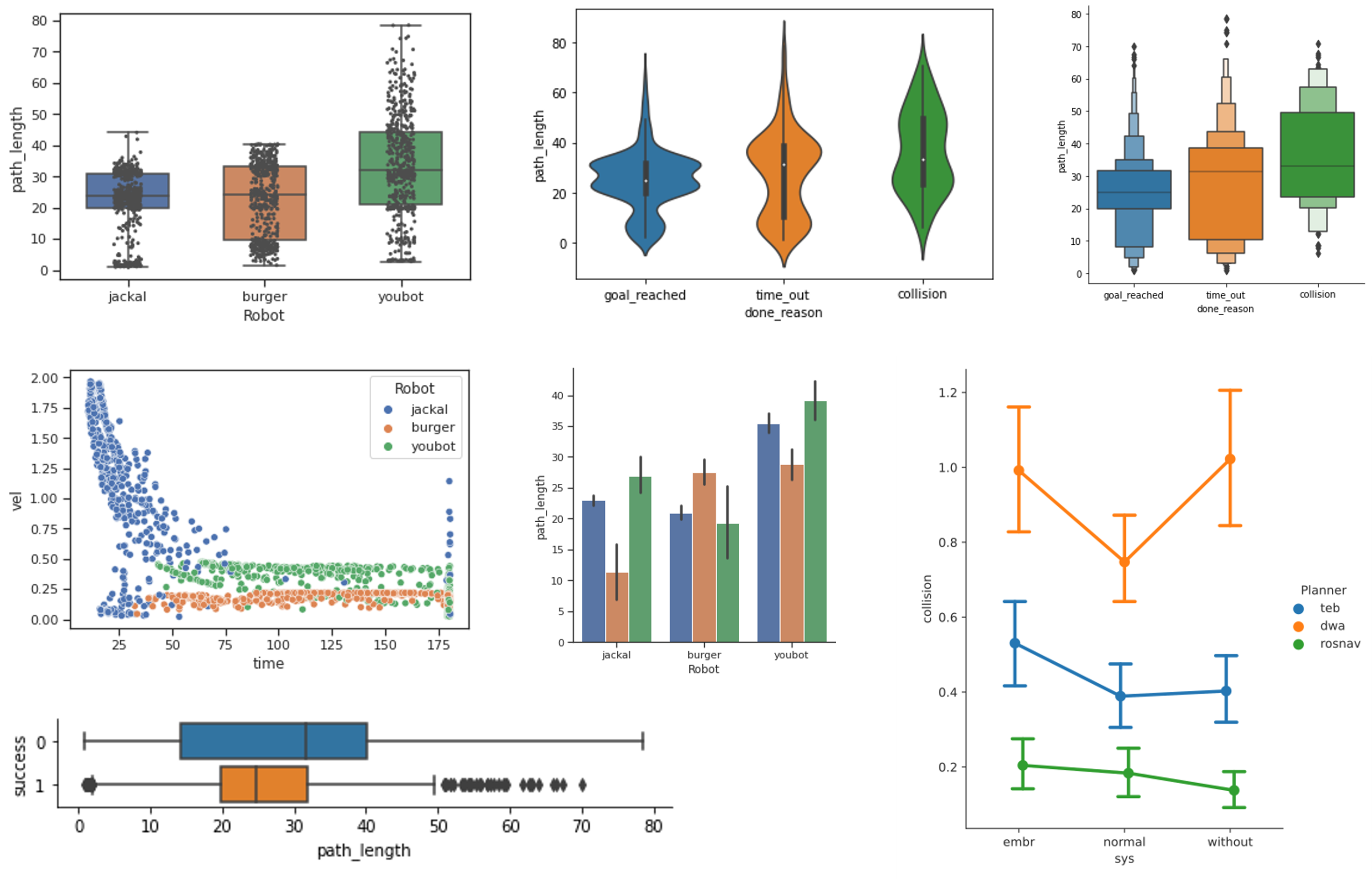}
    \caption{Exemplary Plots created with the evaluation class.}
    \label{quali}
\end{figure}

\noindent
Another important feature of the web-app is the possibility to assess the performance of navigation approaches. It is possible to utilize the application for pure benchmarking of several planners against one another. The user can choose which metrics should be recorded and the implemented data recorder will record only the topics, necessary to calculate the metrics. 
Similar to the training task, starting a new evaluation is coupled with the selection of a high number of different configurations, which can be overwhelming for the user. Therefore, our task editor provides a straightforward user path with specified steps to be taken. From the dashboard, the user can directly select to create a new evaluation task. In the first step, a scenario file is be selected. Although, we highly recommend using scenarios because it allows for a more stable and consistent comparison between multiple evaluation runs, we also offer the possibility to evaluate on random tasks, where obstacles as well as start and goal position of the robot are placed randomly on the map. Therefore, the user has the possibility to toggle between random and scenario mode in the first step. Furthermore, the user can specify the number of evaluation episodes. In the second step, the user selects the robot that should be used for the evaluation. Based on this selection, a planner must be selected in the third step. The user can select between a variety of different planners listed in Table \ref{tab:planners} or select one of the trained DRL agents, which has been created using the web app. In the last step, the user is again prompted to name the evaluation for better identification later on. 
\\\\\noindent
After starting the evaluation the user is redirected to a new page, where the log output of the evaluation is shown. Both, log files and the latest .csv evaluation files can be downloaded. We offer an evaluation class in form of a Jupyter notebook in which pre-existing code and explanations guide the user and allow for individual plotting of the results. This ensure a high degree of customization possibilities for the plots depending on the individual user. Exemplary plots generated with our evaluation class are illustrated in Fig. \ref{quali} In the future an additional live display of the plots inside the web-app is aspired. For the evaluation, a high number of quantitative and qualitative metrics presented in Table \ref{tab:eval-metrics-table} are provided. The required workflow and screens of the web-app are illustrated in Fig. \ref{fig:eval-screens}.

% \subsection{Evaluation and Benchmarking of Approaches}
% Another important feature is the possiblility to assess the performance of the approaches. Therefore the evaluation class of arena-bench is utilized and the planners can be evaluated in up to 14 navigational metrics. The metrics are listed in Table \ref{tab:eval-metrics-table}. 
% 

\begin{table*}[!h]
\centering
\setlength{\tabcolsep}{5pt}
\renewcommand{\arraystretch}{1}
\footnotesize

		\caption{Evaluation metrics}
	\begin{tabular}{p{35mm}cp{42mm}l}
		\hline
		Hyperparameter  & Value & Explanation & Calculation  \\ \hline
		
		\textbf{Safety}& \multicolumn{3}{c}{} \\ \hline

		Collision$^{1,2}$ & -        & Total number of collisions\\
		
		Clearing Dist. $^{2}$ & - & Distance to obstacles \\
		
        \hline
		\textbf{Robustness}& \multicolumn{3}{c}{} \\ 
		\hline
		Success Rate$^{2}$ & \%   & Fraction of episodes with <2 collisions and no timeout\\
		Variance$^{2}$ & - & Variance between all measurements & $v = \sqrt{\sum^{N}_{0}\frac{(s_N-\overline{s}}{N}}$ \\
	    \hline
	    \textbf{Efficiency}& \multicolumn{3}{c}{} \\ 
	    \hline
		Path Length$^{1,2}$  & [$m$]    & Path length in m           \\ 
		Time to reach goal$^{2}$& [$s$] & Time required to reach the goal   \\ 
        \hline 
        \textbf{Smoothness}& \multicolumn{3}{c}{} \\ 
        \hline
		Movement Jerk$^{2}$ & [$\frac{m}{s^3}$] & Average acceleration & $j = |\frac{\Delta a}{\Delta t}|$ , $a = \frac{\Delta v}{\Delta t}$ \\ 
		Roughness$^{2}$ & -   & Inverse trajectory smoothness & $R(\text{x}_i) = \frac{\text{Triangle Area}(\text{x}_i,\text{x}_{i+1},\text{x}_{i+2})}{|\text{x}_{i+2}-\text{x}_{i}|^2}$\\ 
		Angle$^{2}$ & [$\frac{1}{m^2}$] & Angle over the path length & $c_{norm}(\text{x}_i) = \frac{c(\text{x}_i)}{|\text{x}_{i+1}-\text{x}_{i}|+|\text{x}_{i+2}-\text{x}_{i+1}|}$ \\
		\hline
	\end{tabular}
  
	\label{tab:eval-metrics-table}
\end{table*}

\section{Conclusion}
\noindent In this paper, we proposed Arena-Web, a web-based application to develop, train, and benchmark DRL-based navigation planners on different robotic systems. Therefore, the web app offers a highly open accessible user interface for a state-of-the-art mobile robot benchmarking framework. With our appealing and well-designed user interface, we offer even inexperienced users to dive into training and benchmarking of navigational planners without being overwhelmed by the amount of possibilities the app offers for experienced users, therefore, bringing the field of mobile robotics to a very large group and lower the barrier to entry. Arena-Web provides intuitive tools for generating new maps, scenarios, neural network architectures, reward values, and hyperparameters that allows for highly customized training and evaluation processes. The web-app provides a high number of pre-available robots and planners as well as predefined maps and scenarios. The evaluation class includes a variety of navigational metrics. The use of modern software frameworks and a well-structured system design makes the system not only very secure, but also offers a lot of potential for extensions and new features. By allowing the users to make their data publicly available, we offer the possibility of creating a large community where everyone benefits from others' creations. By providing possibilities to download all created data, the user is not limited to any of the functionalities of \textit{Arena-Web}, and can also use small parts of the app separately. Future works aspire to include more visualizations and live videos (e.g. of the training and evaluation runs) inside the web application to improve user understanding and experience even more. As mentioned, ongoing works include the integration of the plotting into the web app, which will also enhance simplicity and usability even more. Furthermore, we currently work on more functionalities, such as robot modeling, and integration of multi-agent benchmarking.
The web-app is free to use and openly available under the link stated in the supplementary materials.

% \noindent In this paper, we proposed Arena-Web, a web-based application to develop, train DRL-based navigation planners on different robotic systems and to benchmark and evaluate them against classic and other learning-based baseline approaches. Arena-Web provides intuitive tools for generating new map, scenario, neural network architectures, reward values and hyperparamters that allows for highly customized training and evaluation processes. The webapplication provides a high number of pre-available robots and planners and its evaluation class includes up to 14 navigational metrics. Additionally, we provide a plotting class for the user to plot the results in an appealing way. The web app is build on top of our previous work arena-bench which is designed in a modular way and can be extended with more functionalities and approaches. Future works aspire to include more visualizations inside the web application to improve user understanding and experience even more. Furthermore, we currently work on more functionalities such as 3D robot modeling, live preview of a training or evaluation run, and an improved design.

%\section*{Acknowledgement}
%We acknowledge help with the production of the spot welds and with the chisel test by Hubert Suwala.

\addtolength{\textheight}{-1cm} 

%%%%%%%%%%%%%%%%%%%%%%%%%%%%%%%%%%%%%%%%%%%%%%%%%%%%%%%%%%%%%%%%%%%%%%%%%%%%%%%%

%%%%%%%%%%%%%%%%%%%%%%%%%%%%%%%%%%%%%%%%%%%%%%%%%%%%%%%%%%%%%%%%%%%%%%%%%%%%%%%%

%%%%%%%%%%%%%%%%%%%%%%%%%%%%%%%%%%%%%%%%%%%%%%%%%%%%%%%%%%%%%%%%%%%%%%%%%%%%%%%%

%%%%%%%%%%%%%%%%%%%%%%%%%%%%%%%%%%%%%%%%%%%%%%%%%%%%%%%%%%%%%%%%%%%%%%%%%%%%%%%%
\typeout{}
\bibliographystyle{IEEEtran}
\bibliography{main}

\end{document}